\documentclass[preprint,12pt]{elsarticle}
\usepackage{booktabs}

\usepackage{amssymb}
\usepackage{amsmath}
\usepackage{hyperref} 
\usepackage{etex}
\usepackage[utf8]{inputenc}
\usepackage[T5]{fontenc}

\usepackage{xcolor}

\journal{arXiv}

\begin{document}

\begin{frontmatter}
\title{HTR-ConvText: Leveraging Convolution and Textual Information for Handwritten Text Recognition}




\author[UIT,VNU]{Pham Thach Thanh Truc} 
\author[UIT,VNU]{Dang Hoai Nam}
\author[UIT,VNU]{\\Huynh Tong Dang Khoa}
\author[UIT,VNU]{Vo Nguyen Le Duy \corref{cor1}}

\affiliation[UIT]{organization={University of Information Technology},
            city={Ho Chi Minh City},
            country={Vietnam}}
            
\affiliation[VNU]{organization={Vietnam National University},
            city={Ho Chi Minh City},
            country={Vietnam}}

            
\cortext[cor1]{Corresponding author. E-mail: duyvnl@uit.edu.vn}

\begin{abstract}

Handwritten Text Recognition remains challenging due to the limited data, high writing style variance, and scripts with complex diacritics.
Existing approaches, though partially address these issues, often struggle to generalize without massive synthetic data.
To address these challenges, we propose HTR-ConvText, a model designed to capture fine-grained, stroke-level local features while preserving global contextual dependencies.
In the feature extraction stage, we integrate a residual Convolutional Neural Network backbone with a MobileViT with Positional Encoding block.
This enables the model to both capture structural patterns and learn subtle writing details.
We then introduce the ConvText encoder, a hybrid architecture combining global context and local features within a hierarchical structure that reduces sequence length for improved efficiency.
Additionally, an auxiliary module injects textual context to mitigate the weakness of Connectionist Temporal Classification.
Evaluation on IAM, READ2016, LAM and HANDS-VNOnDB demonstrate that our approach achieves improved performance and better generalization compared to existing methods, especially in scenarios with limited training samples and high handwriting diversity.

\end{abstract}


\end{frontmatter}



\section{Introduction}
\label{sec1}
Handwritten Text Recognition (HTR) aims to convert images of handwritten text into their corresponding character sequences. Similar to prior studies, we focus on line-level recognition, which provides a practical balance between annotation cost and contextual information richness. However, line-level handwriting recognition remains a challenging problem due to the significant variability in handwriting styles across individuals, writing instruments, and scanning conditions, as well as the linguistic diversity and complexity of character systems, particularly in languages that include diacritics or compound characters.

Traditional approaches based on Convolutional Neural Networks (CNNs) or Recurrent Neural Networks (RNNs) have achieved promising results, yet they often struggle to distinguish fine-grained stroke-level features, such as curvature, slant, pen connectivity, and stroke thickness. Furthermore, these systems predominantly rely on Connectionist Temporal Classification (CTC) for decoding, which inherently lacks textual context modeling capabilities. By predicting characters based solely on visual cues without leveraging semantic dependencies, CTC-based models often exhibit inferior accuracy compared to attention-based methods. More recently, Transformer-based and Vision Transformer (ViT) models have shown strong potential in capturing global dependencies. Nonetheless, they typically lack local inductive bias and require large-scale data to generalize well, making them less effective in low-resource HTR settings, which are common in many practical scenarios.

To address these limitations, we propose HTR-ConvText, a novel architecture designed to combine the strengths of detailed local representations and neighboring spatial context modeling. Specifically, the model employs a CNN-based feature extractor enhanced with an MVP Block at the network’s input stage, enabling it to capture subtle handwriting variations while maintaining strong generalization even on relatively small datasets such as IAM, READ, LAM, and HANDS-VNOnDB. Furthermore, we introduce the ConvText block, a hybrid module that combines Multi-Head Self-Attention (MHSA) for global dependencies with a simplified Depthwise Convolution for local context. This structure is integrated into a U-Net-like hierarchical architecture within the encoder, improving efficiency. We also incorporate a training-only Textual Context Module (TCM) which injects bidirectional textual priors into the visual encoder via an auxiliary loss, which mitigates the linguistic weakness of CTC-based models.

Extensive experiments conducted on multiple benchmark datasets, including IAM, READ, LAM, and  HANDS-VNOnDB, demonstrate that HTR-ConvText consistently outperforms state-of-the-art CNN and Transformer-based methods. Our main contributions are summarized as follows:
\begin{itemize}
\item We propose a compact hybrid CNN–ViT architecture capable of capturing both local and global features, allowing the model to recognize large-scale stroke patterns as well as fine-grained details such as diacritics.

\item We introduce the ConvText encoder, an architecture that unifies local feature and global context modeling. This design further incorporates a hierarchical structure, achieving significant computational efficiency,

\item We develop the Textual Context Module, a training-only component that injects bidirectional textual priors into the visual encoder, addressing the linguistic context weakness of CTC-based models.

\item We conduct thorough cross-lingual evaluations on English, German, Italian and Vietnamese datasets,  showing that our method outperforms existing state-of-the-art approaches. To ensure reproducibility, our implementation along with checkpoints and scripts is openly available at: \href{https://github.com/DAIR-Group/HTR-ConvText}{https://github.com/DAIR-Group/HTR-ConvText}
 
\end{itemize}
\section{Related Work}
\textit{Traditional Approaches to Handwritten Text Recognition.}
Historically, the field of HTR was first approached by statistical methods like Hidden Markov Models (HMMs) \cite{plotz2009, elyacoubi1999}, which relied on handcrafted features. While foundational, HMMs lacked discriminative power and struggled with long-range dependencies.
The emergence of deep learning established a new trend, with Convolutional Recurrent Neural Networks (CRNNs) \cite{puigcerver2017, cojocaru2021, bluche2017gated} becoming the standard. This pipeline uses a CNN for feature extraction, a Bidirectional LSTM (BLSTM) \cite{graves2005} to model sequential context, and is trained with the Connectionist Temporal Classification (CTC) loss \cite{graves2006} for alignment-free, end-to-end recognition.
As an alternative, Encoder-Decoder (Seq2Seq) architectures \cite{michael2019, bluche2017scan, doetsch2016} were introduced, using an attention mechanism \cite{bahdanau2015} to autoregressively generate text. While capable of implicit language modeling, these models are often computationally more expensive than their CTC-based counterparts.
More recently, Fully Convolutional Networks (FCNs) \cite{coquenet2022van, yousef2020origami, coquenet2020fcn} were proposed to replace the sequential bottlenecks of RNNs. Architectures like OrigamiNet \cite{yousef2020origami} and VAN \cite{coquenet2022van} use depth-wise separable convolutions \cite{chollet2017} and gating mechanisms (GateBlocks) \cite{yousef2020gateblocks} to achieve high parallelization and state-of-the-art results. However, this architectural trade-off may inherently limit their ability to model long-range contextual relationships.
\\

\textit{Transformer-based Models for Handwritten Text Recognition.}
The success of the Transformer architecture \cite{vaswani2017} in sequence modeling has motivated its use in HTR as a replacement for RNNs. However, standard Vision Transformer models \cite{dosovitskiy2020} lack the inductive bias of CNNs, necessitating extensive pre-training on large-scale datasets \cite{li2023trocr, fujitake2023dtrocr} to achieve competitive performance.
This dependency is exemplified by data-hungry models like TrOCR \cite{li2023trocr}, which integrates pre-trained BEiT \cite{bao2021beit} and RoBERTa \cite{liu2019roberta} models and requires over 700 million synthetic images, and DTrOCR \cite{fujitake2023dtrocr}, a decoder-only model scaled with nearly 2 billion images. In contrast, research on data-efficient Transformers like DeiT \cite{touvron2021deit} and DropKey \cite{li2023dropkey} has focused on mid-sized datasets, though their direct application to pre-training-free HTR remains challenging.
A third approach integrates Transformers into established HTR pipelines by replacing the BLSTM layers of a CRNN with a Transformer encoder, while retaining the Connectionist Temporal Classification loss. A prominent example is HTR-VT \cite{li2024htrvt}, which uses a CNN feature extractor to feed a Transformer encoder and is optimized with CTC, successfully achieving competitive performance on small datasets without pre-training.
\\

\textit{Hybrid Attention and Convolution Blocks.}
A highly influential trend, originating from Automatic Speech Recognition (ASR), has been to unify self-attention and convolution into a single, hybrid encoder block \cite{gulati2020}. This architecture aims to achieve the "best of both worlds" by augmenting the Transformer's global dependency modeling with the strong local inductive bias of a CNN.
The key precedent is the Conformer architecture \cite{gulati2020}, which established the state-of-the-art in ASR by inserting a 1D convolutional module into the standard Transformer block. This success spurred further research into computational efficiency, leading to models like Squeezeformer \cite{kim2022squeeze}, which uses a temporal U-Net structure, and Zipformer \cite{li2023zipformer}, which introduces a highly efficient scaling Zip-block.
Despite this hybrid architecture becoming the de facto standard in speech processing, its application to the visual and spatial challenges of HTR remains largely unaddressed. This leaves a clear research gap in applying these powerful models to capture the fine-grained, stroke-level features of handwritten text.

\section{Method}
\begin{figure*}[!t]
    \centering
    \includegraphics[width=1.0\linewidth]{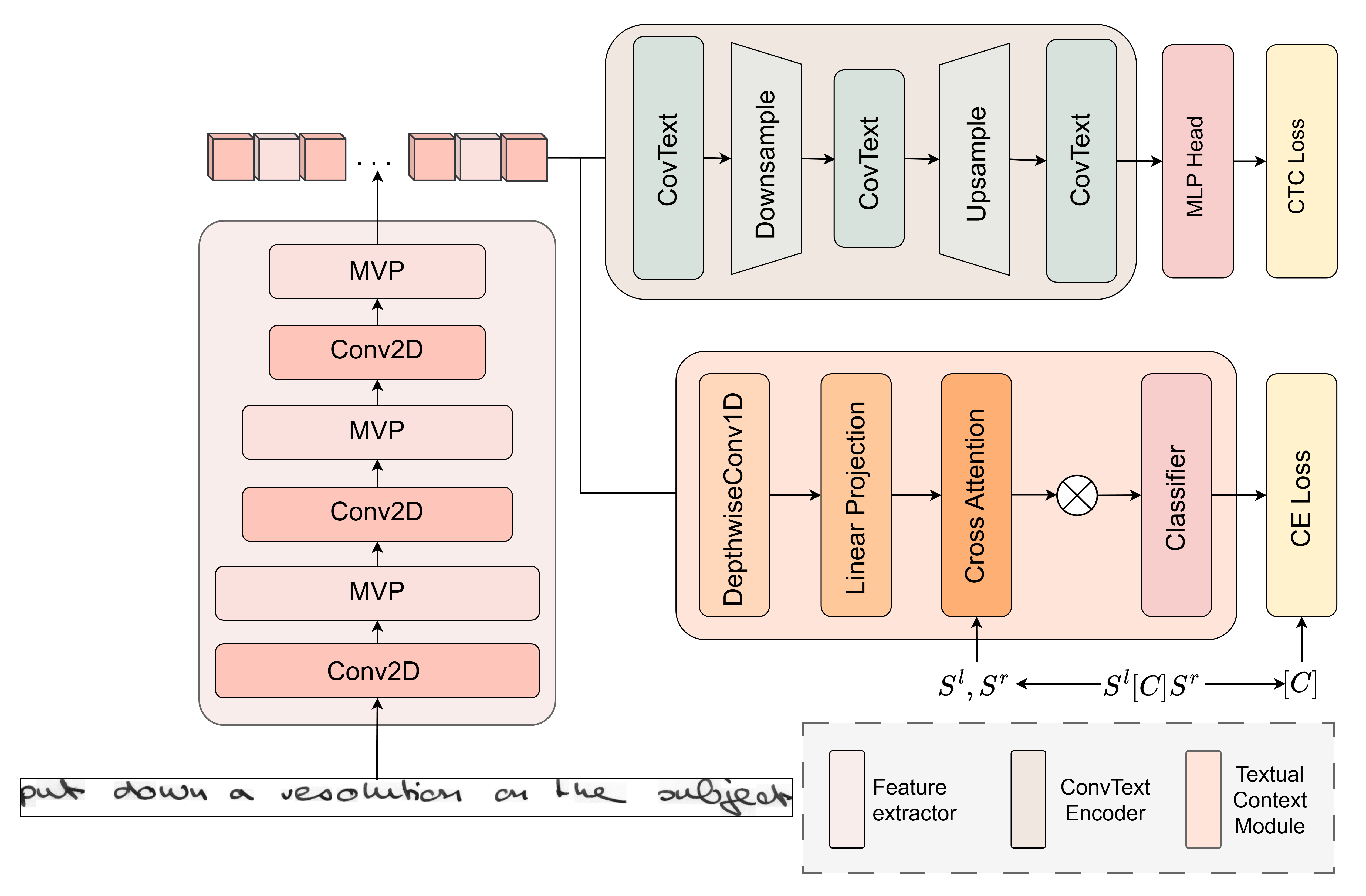}
    \caption{Overview of the HTR-ConvText architecture. The model consists of a CNN-based feature extractor (MVP Block), a hierarchical hybrid encoder (ConvText Encoder), and a training-only Textual Context Module (TCM).}
    \label{fig:architecture}
\end{figure*}

We propose HTR-ConvText, a handwritten text recognition model that combines local convolutional bias with global self-attention, and augments training with a Textual Context Module to help guide training (illustrated in Fig.~\ref{fig:architecture}).

Given a normalized line image $I \in \mathbb{R}^{H \times W}$, a CNN extractor maps $I$ to a 2-D feature map $F \in \mathbb{R}^{H' \times W'}$ with down-sampling factors $(S_h,S_w)$ induced by the convolutional strides. We flatten F row-wise into a sequence $X=\{x_i\}_{i=1}^L$ with $L=\frac{HW}{S_H S_W} ,x_i \in \mathbb{R}$.

A multi-head self-attention and convolutional block then process $X$ into context-aware tokens $\hat{X}$; which are projected to per-step character logits and decoded with a CTC head. We further employ a training-only Textual Context Module that conditions the encoder on left/right textual context and supplies an auxiliary cross-entropy objective. At inference the TCM is omitted, and the model is optimized using the Connectionist Temporal Classification loss and the Cross-entropy Loss.

\subsection{MVP Block}
\begin{figure*}[!t]
    \centering
    \includegraphics[width=1\linewidth]{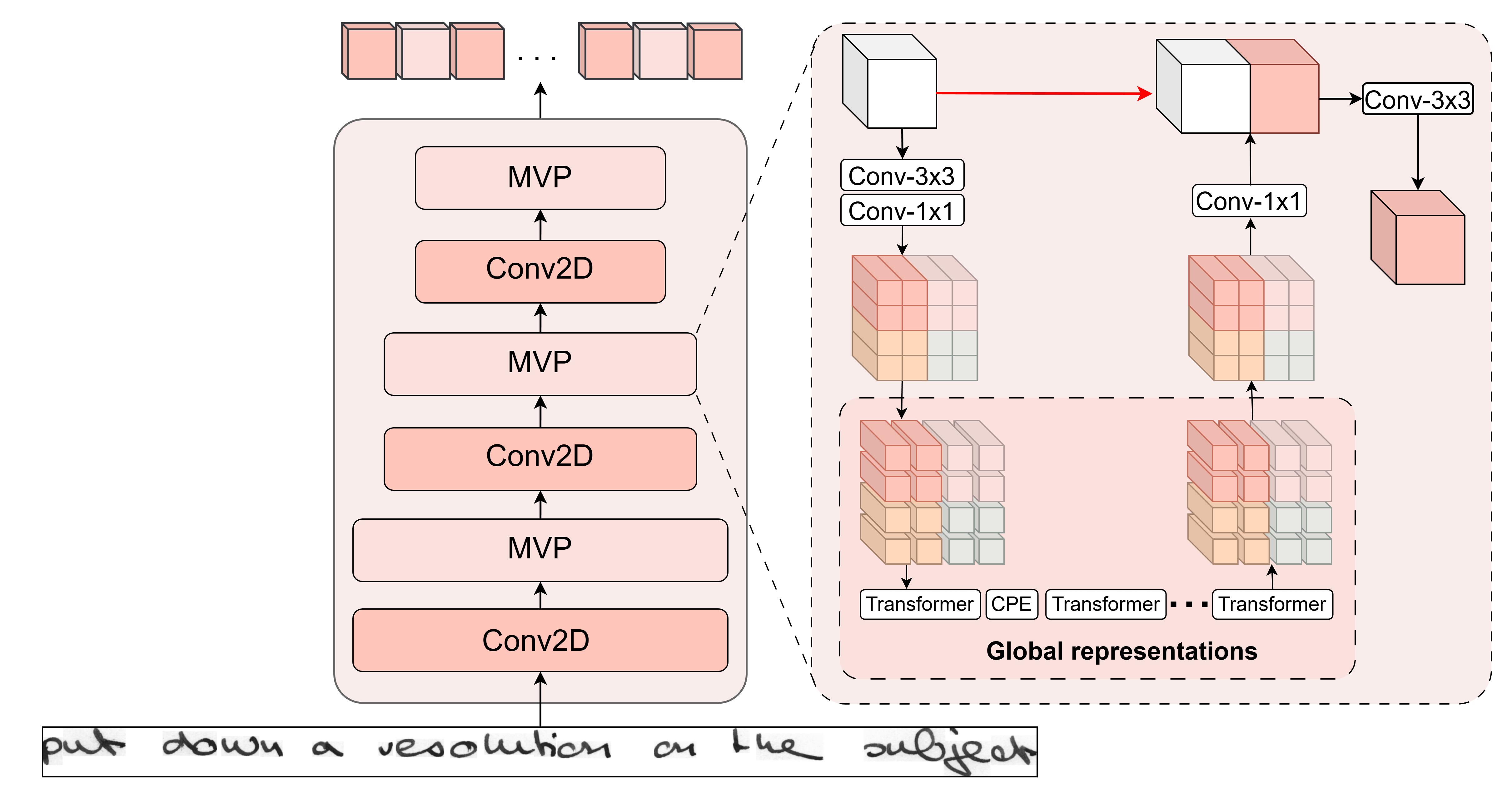}
    \caption{The MVP Block architecture, integrating MobileViT blocks with a Conditional Positional Encoding (CPE) into the ResNet backbone.}
    \label{fig:mvp_block}
\end{figure*}

To extract features that effectively synthesize both local and global (fine-grained) information, we introduce a novel feature extractor, termed the MobileViT \cite{mv} with Positional Encoding (MVP) Block, built upon the synergy of ResNet-18 and the MobileViT Block (MV Block). We maintain the architecture of the initial ResNet-18 almost entirely to fully leverage the inherent strengths of its stacked CNN layers for robust local feature extraction and hierarchical representation learning. Crucially, MobileViT Blocks are strategically integrated, interleaved within the final blocks of the ResNet backbone, as shown in Fig.~\ref{fig:mvp_block}. This intelligent hybrid structure capitalizes on the complementary nature of local features from the CNN and global context captured by the Vision Transformer (ViT) layers, allowing the model to effectively capture the interplay between local details and global scene context, thereby yielding richer and more expressive feature representations. Furthermore, to enable the transformed patches to effectively encode and utilize their relative spatial positions, a mechanism essential for the model's sequence learning and spatial awareness, Positional Encodings (PE) must be incorporated. Since the MobileViT block employs Fold and Unfold operations (which transform 2D feature maps to sequences and back), conventional trigonometric-based PEs are suboptimal for this specific structure. Therefore, we integrate a more robust mechanism: the Conditional Positional Encoding \cite{cpe} (CPE), which is specifically designed to handle features with spatial dimensions, thus facilitating the model's learning process. As demonstrated in our experimental validation, the inclusion of CPE significantly enhances the efficiency and effectiveness of the feature extraction process. In conclusion, the MVP Block represents an efficient and powerful architecture.

\subsection{ConvText Encoder}
Handwritten text lines exhibit strong short-range structure along the writing flow (strokes, ligatures) and long-range dependencies across characters and words. Our encoder therefore interleaves (i) a multi-head self-attention and (ii) a depthwise-separable convolutions along the sequence axis.

We design a hybrid encoder block that couples global self-attention with a lightweight 1-D convolutional module, framed by two half-step feed-forward networks (FFNs). The block adopts post-LayerNorm after every residual update and uses a learnable residual scaling per sub-module, along with stochastic depth for regularization.

\subsubsection{Hybrid ConvText Block} 

\begin{figure*}[!t]
    \centering
    \includegraphics[width=1\linewidth]{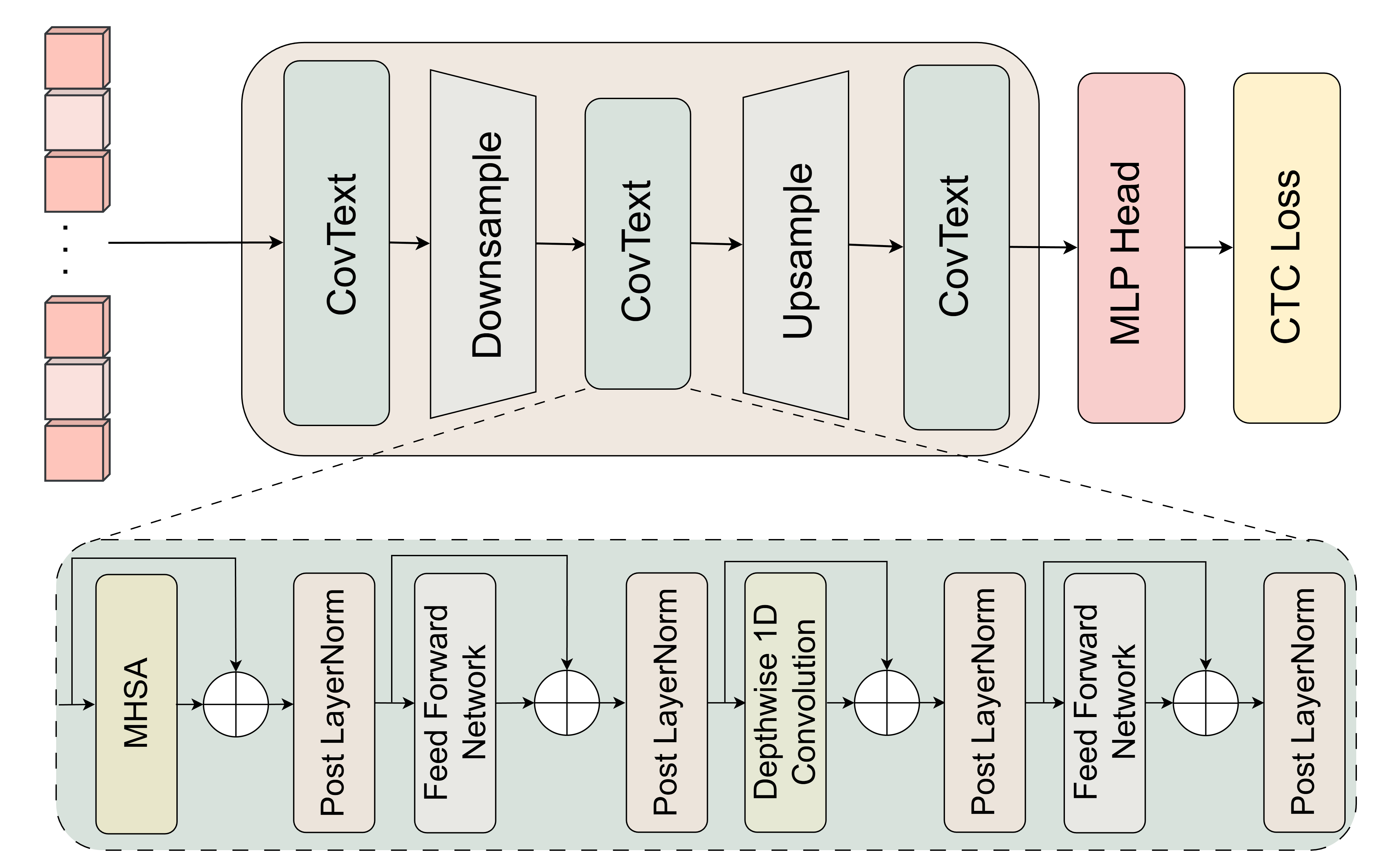}
    \caption{Structure of the Hybrid ConvText Block. It sequentially interleaves Multi-Head Self-Attention (MHSA), a Feed-Forward Network (FFN), a Depthwise Convolution, and a final FFN.}
    \label{fig:convtext}
\end{figure*}

Drawing inspiration from Conformer, we integrate convolution to capture fine-grained local spatial context alongside the global scope of self-attention. However, we diverge from the Conformer's Macaron architecture. Instead, we adopt a sequential structure where a Convolution module and a second Feed-Forward Network are placed after the standard Transformer encoder layers, as illustrated in Fig.~\ref{fig:convtext}. This effectively intertwines local and global feature extraction without the structural redundancy of the Macaron design.

\textit{Relative Positional Encoding.} Unlike HTR-VT, which utilizes fixed sinusoidal positional encodings, we employ relative sinusoidal positional encodings \cite{dai2019transformerxl}. This choice is critical for our architecture; since our encoder employs a U-Net-like hierarchical structure (\S\ref{unet}), the sequence length $L$ changes across stages. Relative encodings ensure the attention mechanism remains robust to these length variations and generalizes better to text lines of arbitrary width.

\textit{Streamlined Convolution Module.} To capture fine-grained local features (such as diacritics or stroke connections), we incorporate a convolutional sub-block. We optimize the standard Conformer design by replacing the computationally expensive Gated Linear Unit (GLU) with a SiLU activation and utilizing a stack of depthwise-separable convolutions. This reduction in parameters improves training stability on smaller datasets without sacrificing the nonlinearity required for local feature modeling.

\subsubsection{Hierarchical U-Net Structure} \label{unet}

The application of standard Transformers to HTR is constrained by the $O(L^2)$ computational cost of the self-attention mechanism, particularly for long text lines. To address this, we organize the ConvText blocks into a hierarchical U-Net structure. 

The encoder begins by processing the input feature sequence $X$ at its original resolution ($L=128$). It then temporally downsamples the sequence by a factor of two ($L'=64$) using a strided 1D convolution. The majority of the depth (4 blocks) is applied in this compressed latent space, effectively quartering the computational cost of the attention mechanism while increasing the effective receptive field. Finally, the sequence is upsampled via nearest-neighbor interpolation and fused with the stored high-resolution features via a residual connection. This allows the model to recover fine-grained spatial details essential for predicting character boundaries.

\subsubsection{Formal Definition}
Mathematically, this means one ConvText block maps $x_0 \in \mathbb{R}^{L \times D}$ to $y \in \mathbb{R}^{L \times D}$ as:
\begin{equation}
\begin{gathered}
    x_1 = \text{LayerNorm}(x_0 + \text{LayerScale}_{\text{attn}}(\text{DP}(\text{MHSA}(x_0)))), \\
    x_2 = \text{LayerNorm}(x_1 + \tfrac{1}{2} \text{LayerScale}_{\text{ffn}1}(\text{DP}(\text{FFN}(x_1)))), \\
    x_3 = \text{LayerNorm}(x_2 + \text{LayerScale}_{\text{conv}}(\text{DP}(\text{Conv}(x_2)))), \\
    y   = \text{LayerNorm}(x_3 + \tfrac{1}{2} \text{LayerScale}_{\text{ffn}2}(\text{DP}(\text{FFN}(x_3)))).
\end{gathered}
\label{eq:convtext_block}
\end{equation}

where $\text{LayerNorm}$ denotes Layer Normalization, and $\text{LayerScale}$ represents a learnable per-channel scaling factor applied to the residual branches. $FFN$, $MHSA$, $DP$ and $Conv$ refer to the Feed-Forward Network, Multi-Head Self-Attention, Drop Path and Convolution modules, respectively.

\subsection{Textual Context Module}

Standard CTC-based recognition operates under a conditional independence assumption, often leading to predictions that lack linguistic coherence. To address this, we introduce a Textual Context Module, a training-only auxiliary branch designed to inject explicit linguistic priors into the visual encoder.

\begin{figure*}[!t]
    \centering
    \includegraphics[width=1\linewidth]{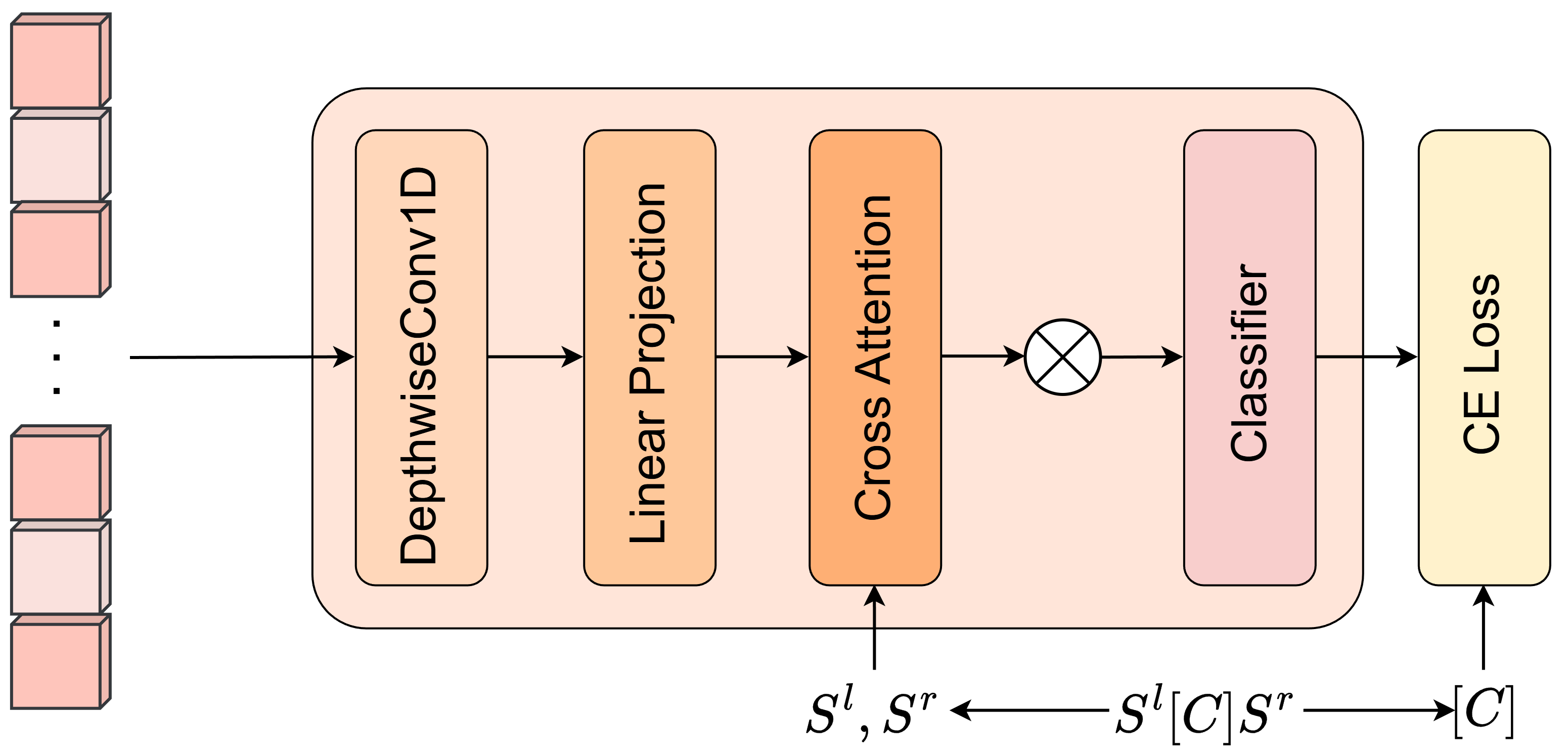}
    \caption{The Textual Context Module (TCM). It generates left and right linguistic context vectors and fuses them with visual features via cross-attention. This module is used only during training.}
    \label{fig:tcm}
\end{figure*}

\subsubsection{Context Formulation}
Let $Y=(c_1, \dots, c_T)$ denote the ground-truth character sequence. For every target character $c_i$, we construct a local bidirectional context comprising a left context window $S_i^l=(c_{i-k}, \dots, c_{i-1})$ and a right context window $S_i^r=(c_{i+1}, \dots, c_{i+k})$, where $k$ is the context window size.

These character sequences are embedded and processed via a lightweight 1D convolutional layer followed by a linear projection to aggregate local textual information. To distinguish the directional provenance of the context, we add learnable directional biases $b_l$ and $b_r$, producing the final textual context vectors $\tilde{q}_{i}^{l}$ and $\tilde{q}_{i}^{r}$:

\begin{equation}
    \tilde{q}_{i}^{l}=\text{LayerNorm}(W_{t}q_{i}^{l}+b_{l}), \quad \tilde{q}_{i}^{r}=\text{LayerNorm}(W_{t}q_{i}^{r}+b_{r})
\end{equation}

where $W_t \in \mathbb{R}^{D \times D}$ is a learnable projection matrix and $\text{LayerNorm}$ denotes Layer Normalization.

To align this linguistic knowledge with the visual representation, we employ a Cross-Attention mechanism. 
The resulting attention-weighted features are fused with the original visual tokens via element-wise multiplication (denoted as $\otimes$ in Fig.~\ref{fig:tcm}) to produce the final conditioned representations, $z_{i}^{l}$ and $z_{i}^{r}$. A shared linear classifier $W_c$ then maps these representations to logits over the vocabulary:

\begin{equation}
    \tilde{y}_{i}^{l}=W_{c}z_{i}^{l}+b_{c}, \quad \tilde{y}_{i}^{r}=W_{c}z_{i}^{r}+b_{c}
\end{equation}

\subsubsection{Auxiliary Objective}
This module creates an auxiliary task: predicting the center character $c_i$ using only the visual features and the surrounding text. This guides the encoder to learn visual representations that are consistent with their linguistic context. 

Crucially, the TCM is utilized only during training. At inference time, the module is discarded, and the model relies solely on the enhanced features of the ConvText encoder. This design allows HTR-ConvText to benefit from linguistic guidance without incurring the latency penalties associated with autoregressive decoders.

\subsubsection{Optimization Objective}
During training, the optimization objective is to minimize the total loss $\mathcal{L}$, which comprises the primary $\mathcal{L}_{CTC}$ from the main encoder and the auxiliary $\mathcal{L}_{TCM}$ from our Textual Context Module.

The $\mathcal{L}_{CTC}$ is the standard Connectionist Temporal Classification loss applied to the final output of the ConvText encoder.

\begin{equation}
\mathcal{L}_{ctc} = CT CLoss(\tilde{Y}_{ctc}, Y)
\label{eq:ctc loss}
\end{equation}

The $\mathcal{L}_{TCM}$ is an auxiliary cross-entropy loss computed from the conditioned representations. We compute the per-position cross-entropy from both left and right directions, denoted as $l_{i}^{l}$ and $l_{i}^{r}$ respectively. These losses are aggregated using a weighted average to handle variable sequence lengths and potential focal 
emphasis:
\begin{equation}
\begin{gathered}
l_{i}^{l}=\text{CE}(\tilde{y}_{i}^{l},c{i}), \quad l_{i}^{r}=\text{CE}(\tilde{y}_{i}^{r},c{i}) \\
\mathcal{L}_{TCM}=\frac{1}{2 \sum_{j=1}^{T} w_{j}} \sum_{i=1}^{T} (l_{i}^{l} + l_{i}^{r}) w_{i
}\end{gathered}
\label{eq:tcm_loss}
\end{equation}

The total objective $\mathcal{L}_{TCM}$ is then calculated as the weighted mean of these bidirectional losses. In this formulation, $w_i$ represents a scalar importance weight (derived from the padding mask, where $w_i=1$ for valid characters and $w_i=0$ for padding tokens). The denominator $2 \sum_{j=1}^{T} w_{j}$ serves as a normalization factor, ensuring that the loss magnitude remains invariant to the sequence length or the batch size.

The full optimization objective is a weighted sum of these two losses:

\begin{equation}
\mathcal{L}=\lambda_{CTC}\mathcal{L}_{CTC}+\lambda_{TCM}\mathcal{L}_{TCM}
\label{eq:loss}
\end{equation}

where $\lambda_{CTC}$ and $\lambda_{TCM}$ are scalar hyperparameters to balance the two objectives. Gradients from $\mathcal{L}_{TCM}$ flow through the cross-attention into the encoder, encouraging text-consistent visual representations. At inference, the TCM is omitted, and decoding uses only the ConvText encoder and the CTC head, incurring no additional latency.

\section{Experiments}

In this section, we evaluate the performance of our model on the line-level text recognition task. The experimental results demonstrate that our approach achieves state-of-the-art performance on the LAM and IAM datasets, while also performing competitively with other leading models on the READ dataset. Notably, our model attains strong results without relying on pre-training, synthetic data, or any additional pre-/post-processing steps.
\\

\subsection{Implementation Details}
The HTR-ConvText encoder consists of 8 stacked ConvText blocks, each containing a multi-head self-attention (MHSA) module with 8 heads, a feed-forward network (FFN) with an expansion ratio of 4.0, and an interleaved depthwise separable convolutional module for local feature aggregation. Each block adopts post-LayerNorm and LayerScale initialization with a scale of 1e-5. The embedding dimension is fixed to 512, and positional encoding follows a fixed 1D relative position bias scheme.
All experiments use images resized to 512×64, with grayscale normalization and a data augmentation pipeline including random dilation/erosion (kernel $\leq$ 2), color jitter, elastic distortion, and geometric transformations applied with a probability of 0.5. The model is optimized using AdamW with a maximum learning rate of 1 × 10$^{-3}$, 1,000 warm-up iterations, and cosine decay scheduling. A weight decay of 0.05 and EMA smoothing (decay = 0.9999) are applied to stabilize training. The batch size is 64 for training and 8 for validation. Trainings are performed on either GPU H100 or A100.

\subsection{Dataset and evaluation metrics}

We evaluated the performance of our model on four handwritten text recognition datasets: IAM \cite{marti2002iam}, READ2016 \cite{sanchez2016read}, LAM \cite{cascianelli2022lam}, and the Vietnamese dataset HANDS-VNOnDB \cite{nguyen2018hands}. Among them, READ2016 and IAM are well-established benchmarks, while LAM represents the largest publicly available line-level dataset. Crucially, we include HANDS-VNOnDB to specifically evaluate the model's robustness in handling complex diacritical marks. Detailed information about these datasets is summarized in Tab.~\ref{table:datasets}. It is important to note that all reported results correspond to the test set, obtained using the model that achieved the best performance on the respective validation sets.

\begin{table}[!t]
    \centering
    \caption{Summary of datasets used for evaluation.}
    \label{table:datasets}
    \begin{tabular}{l|c|c|c|c|c}
        \hline
        Dataset & Training & Validation & Test & Language & Charset \\
        \hline
        \hline
        IAM & 6,482 & 976 & 2,915 & English & 79 \\
        READ2016 & 8,349 & 1,040 & 1,138 & German & 89 \\
        LAM & 19,830 & 2,470 & 3,523 & Italian & 89 \\
        HANDS-VNOnDB & 4,433 & 1,229 & 1,634 & Vietnamese & 161 \\
        \hline
        \hline
    \end{tabular}
\end{table}

\paragraph{LAM Dataset}
The Ludovico Antonio Muratori (LAM) dataset constitutes a large-scale corpus for Handwritten Text Recognition (HTR), focusing on Italian historical manuscripts. As the largest available line-level HTR dataset, LAM comprises 25,823 text lines and features a lexicon of over 23,000 unique words. The standard split utilizes 19,830 lines for training, 2,470 for validation, and 3,523 for testing, with a complete character set of 90 symbols, including '\#'. '\#' is a placeholder to manage transcription ambiguity, illegible text, special symbols, and struck-through words. Being the current largest HTR line-level dataset, LAM serves as a rigorous benchmark for evaluating model performance on long-period historical documents.

\paragraph{READ2016 Dataset}
The READ2016 dataset originates from the ICFHR 2016 competition on HTR, sourced from the Early Modern German Ratsprotokolle collection (READ project). The dataset provides rich annotations at the page, paragraph, and line levels. For line-level HTR, the dataset is partitioned into 8,349 training images, 1,040 validation images, and 1,138 test images, all sharing a total character set size of 89. As a key benchmark, READ2016 is primarily utilized for assessing HTR model capabilities on historical, complex European handwriting.

\paragraph{IAM Dataset}
The IAM dataset is a widely adopted benchmark for offline HTR, specifically focusing on modern English.
It aggregates handwritten transcriptions from the LOB corpus, contributed by 657 distinct writers across 1,539 scanned pages.
Following the common research standard, the line-level subset is split into 6,482 lines for training, 976 lines for validation, and 2,915 lines for testing. All images are provided in grayscale at 300 dpi. Our study utilizes this standard line-level partition of IAM (as detailed in Tab.~\ref{table:datasets}) to compare model performance against established state-of-the-art results for multi-writer modern handwriting recognition.

\paragraph{HANDS-VNOnDB Dataset}
To evaluate our model's robustness on complex diacritics, we utilize the HANDS-VNOnDB dataset, originally collected for the ICFHR 2018 competition on Vietnamese Online Handwritten Text Recognition. Although originating from online pen-tip trajectories, we utilize the rendered offline images to test visual recognition capabilities. Vietnamese script poses a unique challenge due to its high density of stacked diacritical marks (tones), which require fine-grained stroke-level feature extraction to distinguish correctly. The dataset contains handwritten paragraphs produced by university students, offering a diverse range of modern writing styles. We employ a line-level split to assess the model's ability to capture subtle visual details that differentiate tonal markers.

\subsection{Evaluation Metrics}
To evaluate the performance of our model, we employ the standard metrics for handwritten text recognition: Character Error Rate (CER) and Word Error Rate (WER).

Character Error Rate (CER) is defined as the Levenshtein distance between the predicted character sequence and the ground truth. It is calculated as the sum of character substitutions ($S_c$), insertions ($I_c$), and deletions ($D_c$) required to transform the prediction into the ground truth, normalized by the total number of characters in the ground truth ($N_c$). Formally, CER is expressed as:
\begin{equation}
    \text{CER} = \frac{S_c + I_c + D_c}{N_c}
\end{equation}

Word Error Rate (WER) is computed similarly but at the word level. It represents the sum of word substitutions ($S_w$), insertions ($I_w$), and deletions ($D_w$) required to transform the predicted word sequence into the ground truth, divided by the total number of words in the ground truth ($N_w$). It is defined as:
\begin{equation}
    \text{WER} = \frac{S_w + I_w + D_w}{N_w}
\end{equation}

\subsection{Comparison with state-of-the-art approaches}
\begin{table*}[!t]
\centering
\caption{Comparison of Performance (CER, WER) with State-of-the-Art Approaches on the IAM Test Set. Our method achieves the new SOTA with the lowest CER/WER.SOTA results are cited or re-implemented from the HTR-VT paper.}
\label{tab:comparison}
\begin{tabular}{l|c|c|c}
\hline
\textbf{Method} & \textbf{Test CER} & \textbf{Test WER} & \textbf{Param.} \\
\hline
GFCN & 8.0 & 28.6 & 1.4M \\
GFCN$\ast$ & 8.0 & 28.6 & 1.4M \\
CRNN$\ast$ & 7.8 & 27.8 & 18.2M \\
CNN + BLSTM & 8.3 & 24.9 & 9.3M \\
CNN + BLSTM$\ast$ & 7.7 & 26.3 & 9.3M \\
OrigamiNet-12 & 5.3 & - & 39.0M \\
OrigamiNet-12$\ast$ & 6.0 & 22.3 & 39.0M \\
OrigamiNet-18 & 4.8 & - & 77.1M \\
OrigamiNet-18$\ast$ & 6.6 & 24.2 & 77.1M \\
OrigamiNet-24 & 4.8 & - & 115.3M \\
OrigamiNet-24$\ast$ & 6.5 & 23.9 & 115.3M \\
VAN & 5.0 & 16.3 & 2.7M \\
ViT & 32.4 & 68.5 & 37.0M \\
ViT + DropKey & 34.2 & 70.1 & 37.0M \\
DeiT & 32.0 & 68.4 & 6.0M \\
Transformer$^{\S}$ & 4.7 & 15.5 & 54.7M \\
Transformer$\ast$ & 7.6 & 24.5 & 54.7M \\
TrOCR$\ast$ & 7.3 & 37.5 & 385.0M \\
HTR-VT & 4.7 & 14.9 & 53.5M \\

Ours & \textbf{4.0} & \textbf{12.9} & 65.9M \\
\hline
\multicolumn{4}{l}{\footnotesize{$\S$ reports results using extra training data.}} \\
\multicolumn{4}{l}{\footnotesize{$\ast$ indicate re-implementations by LAM.}} \\
\end{tabular}
\end{table*}
\begin{table*}[!t]
\centering
\caption{Performance Comparison on the LAM Dataset. Our approach significantly outperforms competing methods, particularly transformer-based models.SOTA results are cited or re-implemented from the HTR-VT paper.}
\label{tab:comparison_lam}
\begin{tabular}{l|c|c|c}
\hline
\textbf{Method} & \textbf{Test CER} & \textbf{Test WER} & \textbf{Param.} \\
\hline
CNN + BLSTM & 5.8 & 18.4 & 9.3M \\
GFCN$\ast$ & 5.2 & 18.5 & 1.4M \\
CRNN$\ast$ & 3.8 & 12.9 & 18.2M \\
OrigamiNet-12$\ast$ & 3.1 & 11.2 & 39.0M \\
OrigamiNet-18$\ast$ & 3.1 & 11.1 & 77.1M \\
OrigamiNet-24$\ast$ & 3.0 & 11.0 & 115.3M \\
ViT & 6.1 & 19.1 & 37M \\
ViT + DropKey & 5.7 & 16.5 & 37M \\
DeiT & 5.9 & 18.7 & 6M \\
Transformer$^{\S}$ & 10.2 & 22.0 & 54.7M \\
TrOCR$\ast^{\S}$ & 3.6 & 11.6 & 385.0M \\
HTR-VT & 2.8 & 7.4 & 53.5M \\
Ours & \textbf{2.7} & \textbf{7.0} & 65.9M \\
\hline
\multicolumn{4}{l}{\footnotesize{$\S$ reports results using extra training data.}} \\
\multicolumn{4}{l}{\footnotesize{$\ast$ indicates re-implementations by LAM.}} \\
\end{tabular}
\end{table*}

\begin{table*}[!t]
\centering
\caption{Performance Comparison on the READ2016 Test Set. Our approach yields comparable results to the current state-of-the-art on this dataset.SOTA results are cited or re-implemented from the HTR-VT paper.}
\label{tab:comparison_read2016}
\begin{tabular}{l|c|c|c}
\hline
\textbf{Method} & \textbf{Test CER} & \textbf{Test WER} & \textbf{Param.} \\
\hline
CNN + RNN & 5.1 & 21.1 & - \\
CNN + BLSTM & 4.7 & - & - \\
FCN & 4.6 & 21.1 & 19.2M \\
VAN & 4.1 & \textbf{16.3} & 2.7M \\
ViT & 8.5 & 29.6 & 37M \\
ViT + DropKey & 8.1 & 26.4 & 37M \\
DeiT & 8.4 & 28.7 & 6M \\
DAN & 4.1 & 17.6 & 7.6M \\
HTR-VT & 3.9 & 16.5 & 53.5M \\
Ours & \textbf{3.6} & \textbf{15.7} & 65.9M \\
\hline
\end{tabular}
\end{table*}

\begin{table*}[!t]
\centering
\caption{Comparison of Performance (CER, WER) with State-of-the-Art Approaches on the HANDS-VNOnDB Test Set. Our method achieves the new SOTA on the Vietnamese handwriting dataset.}
\label{tab:comparison_vn}
\begin{tabular}{l|c|c|c}
\hline
\textbf{Method} & \textbf{Test CER} & \textbf{Test WER}  & \textbf{Param.} \\
\hline
GFCN & 8.4 & 24.2 & 1.4M\\
CNN + BLSTM & 10.53 & 29.1 & 8.3M \\
OrigamiNet & 7.6 & 22.6 & 39.1M\\
HTR-VT & 4.26 & 11.2  & 53.5M\\
Ours &  \textbf{3.45} & \textbf{8.9} & 65.9M\\
\hline
\end{tabular}
\end{table*}


We present a comprehensive performance comparison of our proposed HTR-ConvText model against state-of-the-art approaches across three widely used public benchmarks: IAM, LAM, and READ2016.
The evaluation is primarily based on the Character Error Rate (CER) and Word Error Rate (WER).
The results are summarized in Tab.~\ref{tab:comparison}, Tab.~\ref{tab:comparison_lam}, and Tab.~\ref{tab:comparison_read2016}.
To evaluate the efficacy of our proposed architecture, we conducted a comprehensive comparison against existing state-of-the-art (SOTA) methods on three standard handwritten text recognition benchmarks: IAM, LAM, and READ2016.
The quantitative results, primarily measured by Character Error Rate (CER) and Word Error Rate (WER), are presented in Tabs.~\ref{tab:comparison}, \ref{tab:comparison_lam}, and \ref{tab:comparison_read2016}.
Across all evaluated scenarios, the proposed HTR-ConvText consistently outperforms competing approaches, establishing new benchmarks for recognition accuracy.

\paragraph{IAM Dataset} 
As shown in Tab.~\ref{tab:comparison}, our model demonstrates significant improvements on the IAM test set.
HTR-ConvText achieves a CER of $4.0\%$ and a WER of $12.9\%$.
Notably, these metrics represent a substantial reduction in error rates compared to the previous best-performing model, HTR-VT, improving by $0.7\%$ in CER and $2.0\%$ in WER.
This confirms the effectiveness of our approach in handling complex handwriting variations.

\paragraph{LAM Dataset}
The results on the LAM dataset (Tab.~\ref{tab:comparison_lam}) further highlight the scalability of our model when trained on larger datasets.
HTR-ConvText sets a new state-of-the-art record with a remarkable CER of $2.7\%$ and a WER of $7.0\%$.
Surpassing the $3.0\%$ CER threshold is a critical milestone, indicating that our model's transcription capability is approaching human-level precision on this specific domain.

\paragraph{READ2016 Dataset}
Consistent with the findings on IAM and LAM, Tab.~\ref{tab:comparison_read2016} evidences the robustness of our method on the READ2016 dataset.
By achieving the lowest error rates (CER: $3.6\%$, WER: $15.7\%$), HTR-ConvText proves its generalizability across different historical handwriting styles and document layouts.

\paragraph{HANDS-VNOnDB Dataset}
As presented in Tab. \ref{tab:comparison_vn}, our proposed HTR-ConvText model also demonstrates strong performance on the HANDS-VNOnDB dataset. With only 4,433 training samples, HTR-ConvText achieves a CER of $3.45\%$ and a WER of $8.9\%$, outperforming the previous state-of-the-art method HTR-VT (CER: $4.26\%$, WER: $11.2\%$). These results indicate that our approach is effective even in low-resource scenarios, and it consistently provides highly accurate recognition of Vietnamese handwritten text.

\subsection{Ablation studies}
To better understand the contribution of each architectural component, we conduct a detailed ablation study that involves both adding and replacing modules within our framework. This experimental setup allows us to examine how performance changes when individual components are introduced or substituted, thereby revealing their specific roles and interactions. By comparing these configurations, we can determine which design choices offer the most significant improvements in CER and WER. These insights play an essential role in guiding the selection of the final architecture for our HTR-ConvText model.
\begin{table*}[!t]
\centering
\caption{
Ablation study of the proposed architecture. Starting from the baseline Vision Transformer with a ResNet backbone (A), each additional component incrementally improves recognition accuracy. The ConvText Block (B) provides significant gains by enhancing local feature modeling, the TCM head (C) further reduces errors through enriched contextual aggregation, and the MVP block (D) achieves the best performance by effectively combining global and local representations.
}
\begin{tabular}{lcc}
\hline
\textbf{Component} & \textbf{CER (\%)} & \textbf{WER (\%)} \\
\hline
Vision Transformer with ResNet backbone (A) & 6.25 & 20.05 \\
(A) + ConvText Block (B) & 4.87 & 15.38 \\
(B) +  TCM head (C) & 4.19 & 13.15 \\
(C) +  MVP block (D) & \textbf{4.02} & \textbf{12.91} \\
\hline
\label{table:component}
\end{tabular}
\end{table*}
\subsubsection{Impact of the components}
Table~\ref{table:component} summarizes the ablation study, illustrating the contribution of each architectural component.  
Starting from the baseline Vision Transformer with a ResNet backbone (A), the model achieves a CER of $6.25\%$ and a WER of $20.05\%$. Integrating the ConvText Block (B) significantly boosts performance, confirming the necessity of enhanced local feature modeling for handwriting recognition. Adding the TCM head (C) further decreases both CER and WER, demonstrating its effectiveness in aggregating complementary contextual information. Introducing the MVP block (D) yields the best overall results, showing that the combination of MobileViT-style processing and positional priors provides additional robustness to variations in handwriting style. Overall, the results validate that each proposed component contributes positively and cumulatively to the final recognition performance.

\begin{table*}[!t]
\centering
\caption{Ablation study on Transformer layers and heads.}
\label{table:layers and heads}
\begin{tabular}{|cc|c|c|c|c|}
\hline
\multicolumn{2}{|c|}{} & \multicolumn{4}{c|}{IAM [33]} \\
\hline
Layers & Heads & Val CER & Val WER & Test CER & Test WER \\
\hline
12 & 8 & 2.71 & 8.94 & 4.11 & 13.02 \\
10 & 8 & 2.68 & 9.08 & 4.19 & 13.33 \\
\textbf{8} & 8 & 2.71 & 9.04 & 4.06 & 12.93 \\
6 & 8 & 2.72 & 9.08 & 4.08 & 13.08 \\
4 & 6 & 2.77 & 9.28 & 4.23 & 13.62 \\
\hline
Layers & Heads & Val CER & Val WER & Test CER & Test WER \\
\hline
8 & \textbf{12} & 2.71 & 8.94 & 4.02 & 12.92 \\
8 & 8 & 2.71 & 9.04 & 4.06 & 12.93 \\
8 & 6 & 2.71 & 9.07 & 4.08 & 13.1\\
8 & 4 & 3.00 & 9.95 & 4.59 & 14.61 \\
\hline
\end{tabular}
\end{table*}

\subsubsection{Visualization of attention maps}

\begin{figure*}[!t]
    \centering
    \includegraphics[width=1.0\linewidth]{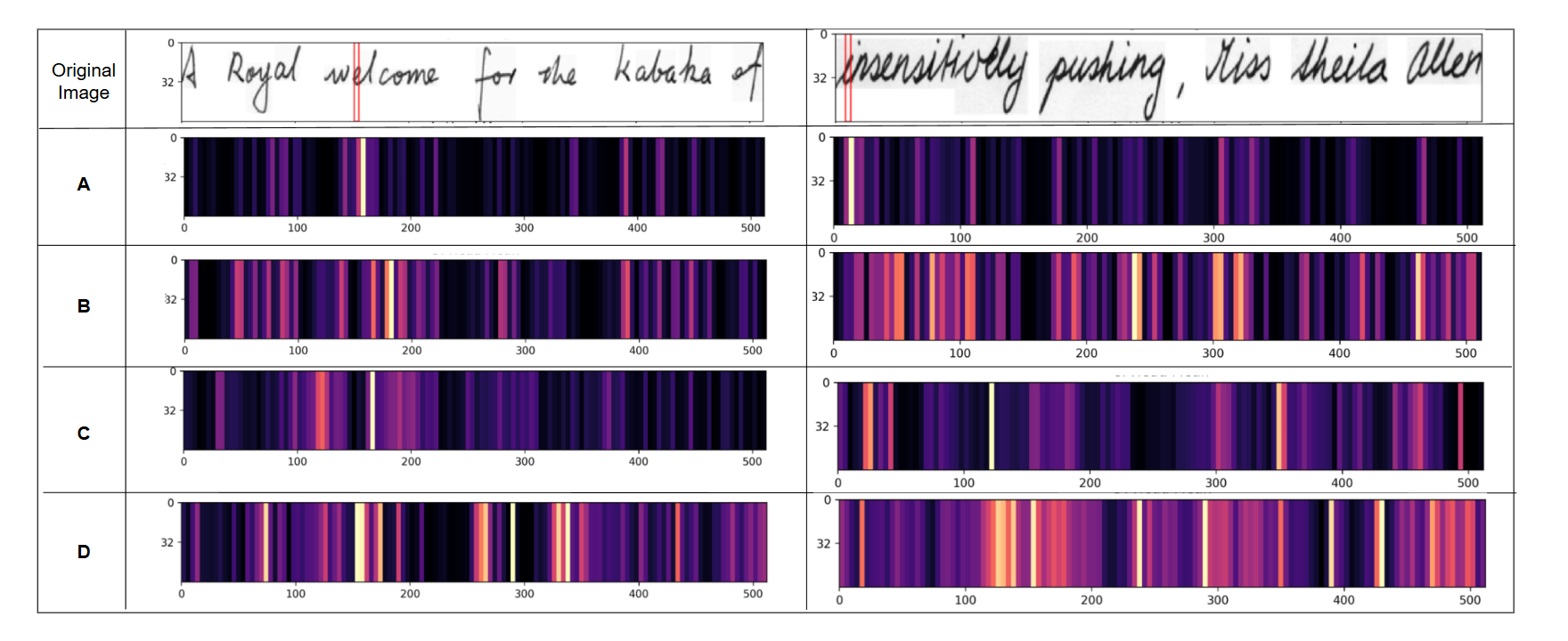}
    \caption{Visualization of attention maps for the four model configurations (A--D). Brighter colors indicate stronger attention. The red boxes on the input images mark the tokens selected for visualization.}
    \label{fig:attention_map}
\end{figure*}

Figure~\ref{fig:attention_map} illustrates the averaged self-attention distributions for the four model configurations (A--D). To analyze how each architectural component affects contextual reasoning, we select a specific token corresponding to the character highlighted by the red box in the input images. From the last encoder layer, we extract the self-attention weights, average them across all heads, and visualize the resulting $1 \times 128$ attention vector as a heatmap, where brighter colors indicate stronger attention.

\textit{Configuration A.}
As shown in Figure~\ref{fig:attention_map}, the baseline model exhibits a highly localized attention pattern. The selected token focuses almost exclusively on itself, suggesting that the baseline primarily relies on the visual features of the target patch and captures limited contextual information.

\textit{Configuration B.}
With the addition of the ConvText Block, the attention begins to extend to neighboring tokens. The model shows increased responsiveness to surrounding patches, demonstrating that ConvText strengthens local feature interactions and encourages the use of short-range context.

\textit{Configuration C.}
After incorporating the TCM head, the attention further broadens. The selected token now attends not only to local neighbors but also to additional contextually relevant regions. This indicates that the TCM head enhances the aggregation of complementary contextual cues.

\textit{Configuration D.}
The final configuration achieves the most balanced attention distribution. As visible in Fig.~\ref{fig:attention_map}, the selected token maintains a strong focus on the target character while simultaneously attending to a broader set of meaningful surrounding tokens. This behavior highlights the effectiveness of the MVP block in integrating global and local features, leading to more robust recognition under handwriting variations.

\subsubsection{Impact of different hyperparameters} 
We investigate the impact of different ConvText encoder layers and attention heads on the model's performance.
The results are illustrated in Tab.~\ref{table:layers and heads}. On the IAM dataset, setting the number of layers to 8 achieved the best balance of Validation CER and WER. While increasing the number of attention heads to 12 resulted in a marginal performance gain (achieving the lowest Test CER of $4.02\%$), it came at the cost of significantly larger file size and increased training time. As we strive for an architecture that is both lightweight and efficient, we prioritized computational economy over marginal accuracy gains. Consequently, we selected 8 encoder layers and 8 attention heads as the optimal configuration, and we employed this set of parameters across all our experiments to maintain consistency.
\begin{table*}[!t]
\centering
\caption{Ablation study on the effect of different positional embeddings (PEs) on the performance of the MVP block. We report CER and WER on the IAM dataset.}
\label{table:pe}
\begin{tabular}{lcc}
\hline
\textbf{Component} & \textbf{CER (\%)} & \textbf{WER (\%)} \\
\hline
Ours with MV block & 4.12 & 13.09 \\
Ours with LPE &  4.06 & 12.97 \\
Ours with RPE & 4.02 & 12.99 \\
Ours with CPE & \textbf{4.02} & \textbf{12.91} \\
\hline
\end{tabular}
\end{table*}

\subsubsection{Impact of different positional embeddings}
Table~\ref{table:pe} summarizes the ablation results of different positional embeddings (PEs) applied to the MVP block. All variants outperform the baseline MV block (CER $4.12\%$, WER $13.09\%$), confirming the importance of positional information. The Learnable Positional Embedding (LPE) offers a modest improvement, while the Relative Positional Embedding \cite{rpe} (RPE) provides a similar level of enhancement. Notably, the Conditional Positional Encoding (CPE) achieves the most favorable balance between CER and WER (CER $4.02\%$, WER $12.91\%$), demonstrating its strong ability to encode spatial relationships within handwriting patterns. These results support our choice of CPE as the positional embedding strategy for the MVP block, as it consistently yields the most reliable performance gains.

\subsection{Qualitative Results}

\begin{figure*}[!t]
    \centering
    \includegraphics[width=\linewidth]{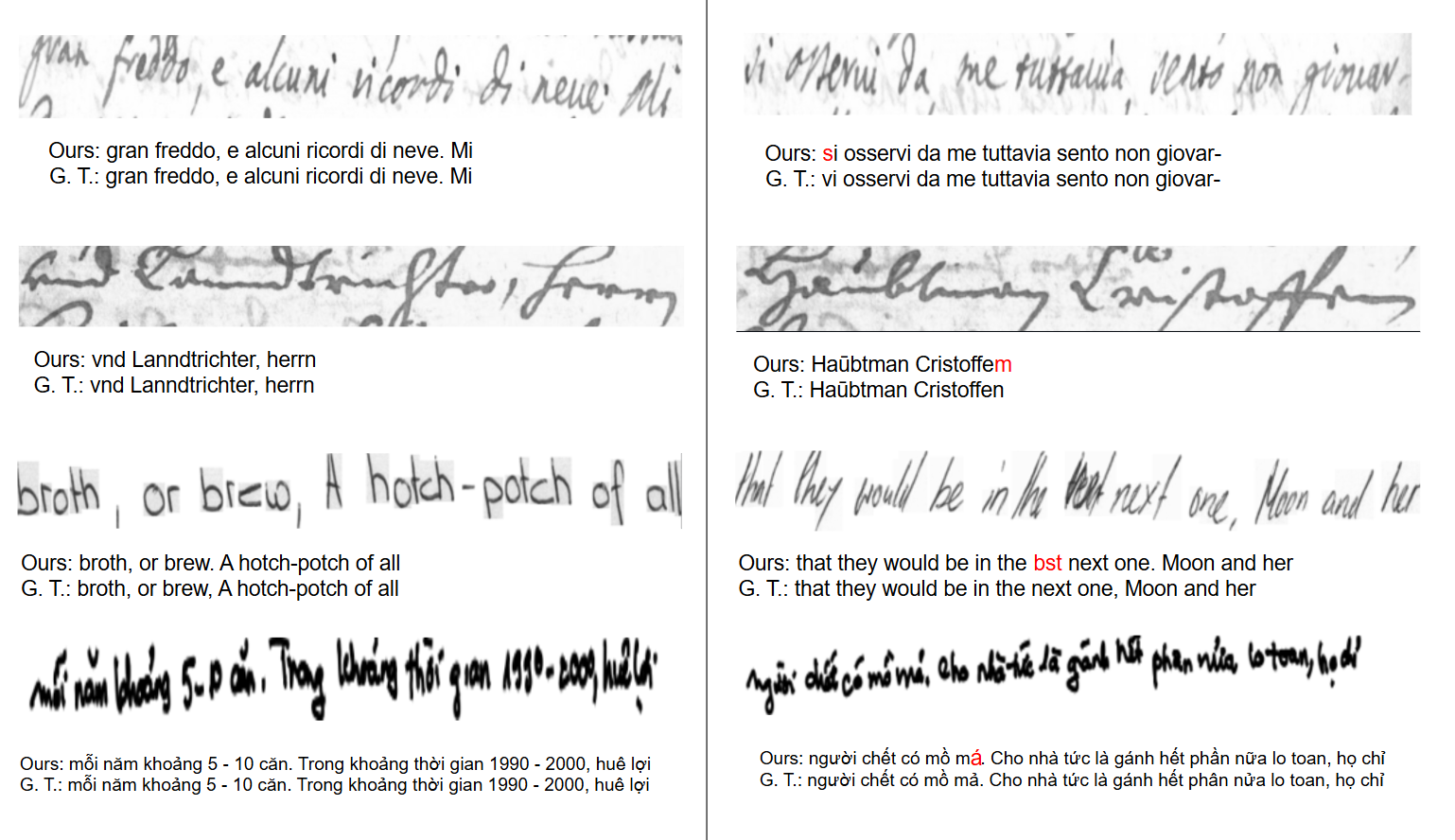}
    \caption{
        Qualitative examples from four datasets (LAM, READ 2016, IAM, and HANDS-VNOnDB). 
        For each dataset, the left column presents correctly recognized samples, 
        while the right column shows typical failure cases. 
        These examples illustrate the primary challenges that still lead to misrecognition, 
        including irregular character shapes, continuous cursive ligatures, text degradation, 
        and compact Vietnamese diacritics.
    }
    \label{fig:qualitative}
\end{figure*}

To further assess the behavior of our model beyond quantitative metrics, 
Fig.~\ref{fig:qualitative} provides qualitative examples drawn from four benchmark datasets: 
LAM, READ 2016, IAM, and HANDS-VNOnDB. 
For each dataset, the left column shows successful predictions, while the right column illustrates representative failure cases.

\textit{LAM (Row 1)s}  
Correct predictions occur in relatively clean handwriting, whereas misrecognition typically arises from highly irregular character shapes. 
In the shown example, the letter ``v'' is written with an unusually tall ascender, causing ambiguity and leading to an incorrect prediction.

\textit{READ 2016 (Row 2).}  
This dataset contains fully cursive writing, and errors frequently result from characters being tightly connected. 
In the failure case, the model confuses ``n'' with ``m'' due to elongated ligatures that merge adjacent strokes.

\textit{IAM (Row 3).}  
Mistakes often stem from visual obstructions or text degradation. 
The example illustrates a word partially crossed out, where the overlaid stroke distorts the underlying characters, causing the system to misread the text.

\textit{HANDS-VNOnDB (Row 4).}  
Vietnamese tonal marks are compact and closely attached to base characters. 
Errors usually occur when diacritics are written too narrowly or faintly. 
In the shown case, the model incorrectly predicts the tone mark because the vowel–diacritic spacing is extremely tight.

Our qualitative evaluation reveals that residual recognition errors are predominantly attributable to intrinsic challenges inherent in handwritten data, including substantial handwriting variability, stroke distortions, character overlap, and fine-grained diacritic compaction. These observations strongly suggest that the core challenges lie within the data characteristics, rather than architectural limitations. Consequently, future research should focus on enhancing model robustness and data invariance to address these persistent difficulties

\section{Conclusion}
In this work, we successfully introduced HTR-ConvText, a novel and compact hybrid CNN–ViT architecture that effectively addresses the intrinsic challenges of Handwritten Text Recognition (HTR) by judiciously combining detailed local feature extraction with global contextual modeling. Our primary contributions include the proposal of a compact hybrid design, the efficient ConvText encoder with a hierarchical structure, and the Textual Context Module (TCM), which mitigates the inherent linguistic context weaknesses of CTC-based systems. Through extensive cross-lingual evaluations conducted on four diverse benchmark datasets (IAM, READ, LAM, and HANDS-VNOnDB), HTR-ConvText consistently demonstrated superior performance, achieving new state-of-the-art results when compared against leading CNN and Transformer-based methodologies, thereby affirming its robustness and strong generalization capabilities across multiple languages. Looking forward, our analysis of residual errors suggests that future efforts should concentrate on enhancing model robustness against challenging intrinsic data properties, such as severe stroke distortion and character overlap. Furthermore, exploring the feasibility of integrating the TCM's linguistic modeling capabilities into an inference-time component remains a promising avenue for improving real-world HTR application efficiency.

\bibliographystyle{elsarticle-num}
\bibliography{references}

@misc{rpe,
      title={Self-Attention with Relative Position Representations}, 
      author={Peter Shaw and Jakob Uszkoreit and Ashish Vaswani},
      year={2018},
      eprint={1803.02155},
      archivePrefix={arXiv},
      primaryClass={cs.CL},
      url={https://arxiv.org/abs/1803.02155}, 
}

@misc{cpe,
      title={Conditional Positional Encodings for Vision Transformers}, 
      author={Xiangxiang Chu and Zhi Tian and Bo Zhang and Xinlong Wang and Chunhua Shen},
      year={2023},
      eprint={2102.10882},
      archivePrefix={arXiv},
      primaryClass={cs.CV},
      url={https://arxiv.org/abs/2102.10882}, 
}

@misc{mv,
      title={MobileViT: Light-weight, General-purpose, and Mobile-friendly Vision Transformer}, 
      author={Sachin Mehta and Mohammad Rastegari},
      year={2022},
      eprint={2110.02178},
      archivePrefix={arXiv},
      primaryClass={cs.CV},
      url={https://arxiv.org/abs/2110.02178}, 
}

@article{plotz2009,
  author  = {Pl{\"o}tz, T. and Fink, G. A.},
  title   = {Markov models for offline handwriting recognition: a survey},
  journal = {International Journal on Document Analysis and Recognition (IJDAR)},
  volume  = {12},
  number  = {4},
  pages   = {269--298},
  year    = {2009}
}

@article{elyacoubi1999,
  author  = {El-Yacoubi, M. A. and Gilloux, M. and Sabourin, R. and Suen, C. Y.},
  title   = {An {HMM}-based approach for off-line unconstrained handwritten word modeling and recognition},
  journal = {IEEE Transactions on Pattern Analysis and Machine Intelligence (TPAMI)},
  volume  = {21},
  number  = {8},
  pages   = {752--760},
  year    = {1999}
}

@inproceedings{puigcerver2017,
  author    = {Puigcerver, Joan},
  title     = {Are multidimensional recurrent layers really necessary for handwritten text recognition?},
  booktitle = {2017 14th IAPR International Conference on Document Analysis and Recognition (ICDAR)},
  volume    = {1},
  pages     = {67--72},
  year      = {2017},
  organization = {IEEE}
}

@inproceedings{cojocaru2021,
  author    = {Cojocaru, I. and Cascianelli, S. and Baraldi, L. and Corsini, M. and Cucchiara, R.},
  title     = {Watch your strokes: Improving handwritten text recognition with deformable convolutions},
  booktitle = {2020 25th International Conference on Pattern Recognition (ICPR)},
  pages     = {6096--6103},
  year      = {2021},
  organization = {IEEE}
}

@inproceedings{bluche2017gated,
  author    = {Bluche, T. and Messina, R.},
  title     = {Gated convolutional recurrent neural networks for multilingual handwriting recognition},
  booktitle = {2017 14th IAPR International Conference on Document Analysis and Recognition (ICDAR)},
  volume    = {1},
  pages     = {646--651},
  year      = {2017},
  organization = {IEEE}
}

@article{graves2005,
  author  = {Graves, Alex and Schmidhuber, J{\"u}rgen},
  title   = {Framewise phoneme classification with bidirectional {LSTM} and other neural network architectures},
  journal = {Neural Networks},
  volume  = {18},
  number  = {5-6},
  pages   = {602--610},
  year    = {2005}
}

@inproceedings{graves2006,
  author    = {Graves, Alex and Fern{\'a}ndez, Santiago and Gomez, Faustino and Schmidhuber, J{\"u}rgen},
  title     = {Connectionist temporal classification: labelling unsegmented sequence data with recurrent neural networks},
  booktitle = {Proceedings of the 23rd International Conference on Machine Learning},
  pages     = {369--376},
  year      = {2006}
}

@inproceedings{michael2019,
  author    = {Michael, J. and Labahn, R. and Gr{\"u}ning, T. and Z{\"o}llner, J.},
  title     = {Evaluating sequence-to-sequence models for handwritten text recognition},
  booktitle = {2019 International Conference on Document Analysis and Recognition (ICDAR)},
  pages     = {1286--1293},
  year      = {2019},
  organization = {IEEE}
}

@inproceedings{bluche2017scan,
  author    = {Bluche, T. and Louradour, J. and Messina, R.},
  title     = {Scan, attend and read: End-to-end handwritten paragraph recognition with {MDLSTM} attention},
  booktitle = {2017 14th IAPR International Conference on Document Analysis and Recognition (ICDAR)},
  volume    = {1},
  pages     = {1050--1055},
  year      = {2017},
  organization = {IEEE}
}

@inproceedings{doetsch2016,
  author    = {Doetsch, P. and Zeyer, A. and Ney, H.},
  title     = {Bidirectional decoder networks for attention-based end-to-end offline handwriting recognition},
  booktitle = {2016 15th International Conference on Frontiers in Handwriting Recognition (ICFHR)},
  pages     = {361--366},
  year      = {2016},
  organization = {IEEE}
}

@article{coquenet2022van,
  author  = {Coquenet, D. and Chatelain, C. and Paquet, T.},
  title   = {End-to-end handwritten paragraph text recognition using a vertical attention network},
  journal = {IEEE Transactions on Pattern Analysis and Machine Intelligence},
  volume  = {45},
  number  = {1},
  pages   = {508--524},
  year    = {2022}
}

@inproceedings{yousef2020origami,
  author    = {Yousef, M. and Bishop, T. E.},
  title     = {{OrigamiNet}: weakly-supervised, segmentation-free, one-step, full page text recognition by learning to unfold},
  booktitle = {Proceedings of the IEEE/CVF Conference on Computer Vision and Pattern Recognition},
  pages     = {14710--14719},
  year      = {2020}
}

@inproceedings{coquenet2020fcn,
  author    = {Coquenet, D. and Chatelain, C. and Paquet, T.},
  title     = {Recurrence-free unconstrained handwritten text recognition using gated fully convolutional network},
  booktitle = {2020 17th International Conference on Frontiers in Handwriting Recognition (ICFHR)},
  pages     = {19--24},
  year      = {2020},
  organization = {IEEE}
}

@inproceedings{vaswani2017,
  author    = {Vaswani, Ashish and Shazeer, Noam and Parmar, Niki and Uszkoreit, Jakob and Jones, Llion and Gomez, Aidan N. and Kaiser, {\L}ukasz and Polosukhin, Illia},
  title     = {Attention is all you need},
  booktitle = {Advances in Neural Information Processing Systems},
  volume    = {30},
  year      = {2017}
}

@article{dosovitskiy2020,
  author  = {Dosovitskiy, Alexey and Beyer, Lucas and Kolesnikov, Alexander and Weissenborn, Dirk and Zhai, Xiaohua and Unterthiner, Thomas and Dehghani, Mostafa and Minderer, Matthias and Heigold, Georg and Gelly, Sylvain and others},
  title   = {An image is worth 16x16 words: Transformers for image recognition at scale},
  journal = {arXiv preprint arXiv:2010.11929},
  year    = {2020}
}

@inproceedings{li2023trocr,
  author    = {Li, Minghao and Lv, Tengchao and Chen, Jingye and Cui, Lei and Lu, Yubo and Florencio, Dinei and Zhang, Cha and Li, Zhoujun and Wei, Furu},
  title     = {{TrOCR}: Transformer-based optical character recognition with pre-trained models},
  booktitle = {Proceedings of the AAAI Conference on Artificial Intelligence},
  volume    = {37},
  pages     = {13094--13102},
  year      = {2023}
}

@article{bao2021beit,
  author  = {Bao, Hangbo and Dong, Li and Piao, Songhao and Wei, Furu},
  title   = {{BEiT}: {BERT} pre-training of image transformers},
  journal = {arXiv preprint arXiv:2106.08254},
  year    = {2021}
}

@article{liu2019roberta,
  author  = {Liu, Yinhan and Ott, Myle and Goyal, Naman and Du, Jingfei and Joshi, Mandar and Chen, Danqi and Levy, Omer and Lewis, Mike and Zettlemoyer, Luke and Stoyanov, Veselin},
  title   = {{RoBERTa}: A robustly optimized {BERT} pretraining approach},
  journal = {arXiv preprint arXiv:1907.11692},
  year    = {2019}
}

@article{fujitake2023dtrocr,
  author  = {Fujitake, Masato},
  title   = {{DTrOCR}: Decoder-only transformer for optical character recognition},
  journal = {arXiv preprint arXiv:2308.15996},
  year    = {2023}
}

@inproceedings{touvron2021deit,
  author    = {Touvron, Hugo and Cord, Matthieu and Douze, Matthijs and Massa, Francisco and Sablayrolles, Alexandre and J{\'e}gou, Herv{\'e}},
  title     = {Training data-efficient image transformers \& distillation through attention},
  booktitle = {International Conference on Machine Learning},
  pages     = {10347--10357},
  year      = {2021},
  organization = {PMLR}
}

@inproceedings{li2023dropkey,
  author    = {Li, Bonan and Hu, Yinqiao and Nie, Xiwen and Han, Congying and Jiang, Xiangjian and Guo, Tiande and Liu, Luoqi},
  title     = {{DropKey} for vision transformer},
  booktitle = {Proceedings of the IEEE/CVF Conference on Computer Vision and Pattern Recognition},
  pages     = {22700--22709},
  year      = {2023}
}

@article{li2024htrvt,
  author  = {Li, Yuting and Chen, Dexiong and Tang, Tinglong and Shen, Xi},
  title   = {{HTR-VT}: Handwritten text recognition with vision transformer},
  journal = {Pattern Recognition},
  year    = {2024}
}

@inproceedings{bahdanau2015,
  author    = {Bahdanau, Dzmitry and Cho, Kyunghyun and Bengio, Yoshua},
  title     = {Neural machine translation by jointly learning to align and translate},
  booktitle = {International Conference on Learning Representations (ICLR)},
  year      = {2015}
}

@inproceedings{chollet2017,
  author    = {Chollet, Fran{\c{c}}ois},
  title     = {Xception: Deep learning with depthwise separable convolutions},
  booktitle = {Proceedings of the IEEE Conference on Computer Vision and Pattern Recognition},
  pages     = {1251--1258},
  year      = {2017}
}

@article{yousef2020gateblocks,
  author  = {Yousef, Mohamed and Hussain, K. F. and Mohammed, U. S.},
  title   = {Accurate, data-efficient, unconstrained text recognition with convolutional neural networks},
  journal = {Pattern Recognition},
  volume  = {108},
  pages   = {107482},
  year    = {2020}
}

@article{marti2002iam,
  author  = {Marti, U.-V. and Bunke, Horst},
  title   = {The {IAM}-database: an english sentence database for offline handwriting recognition},
  journal = {International Journal on Document Analysis and Recognition},
  volume  = {5},
  pages   = {39--46},
  year    = {2002}
}

@inproceedings{sanchez2016read,
  author    = {Sanchez, Joan Andreu and Romero, Veronica and Toselli, Alejandro H and Vidal, Enrique},
  title     = {{ICFHR2016} competition on handwritten text recognition on the {READ} dataset},
  booktitle = {2016 15th International Conference on Frontiers in Handwriting Recognition (ICFHR)},
  pages     = {630--635},
  year      = {2016},
  organization = {IEEE}
}

@inproceedings{cascianelli2022lam,
  author    = {Cascianelli, Silvia and Pippi, Vittorio and Maarand, Martin and Cornia, Marcella and Baraldi, Lorenzo and Kermorvant, Christopher and Cucchiara, Rita},
  title     = {The {LAM} dataset: A novel benchmark for line-level handwritten text recognition},
  booktitle = {2022 26th International Conference on Pattern Recognition (ICPR)},
  pages     = {1506--1513},
  year      = {2022},
  organization = {IEEE}
}

@inproceedings{nguyen2018hands,
  author    = {Nguyen, K. C. and Nguyen, C. T. and Duong, T. V. and Nakagawa, M.},
  title     = {{HANDS-VNOnDB}: A Vietnamese online handwritten database and its application to online Vietnamese handwritten text recognition},
  booktitle = {2018 16th International Conference on Frontiers in Handwriting Recognition (ICFHR)},
  pages     = {465--470},
  year      = {2018},
  organization = {IEEE}
}

@inproceedings{gulati2020,
  author    = {Gulati, Anmol and Qin, James and Chiu, Chung-Cheng and Parmar, Niki and Zhang, Yu and Yu, Jiahui and Han, Wei and Wang, Shibo and Zhang, Zhengdong and Wu, Yonghui and Pang, Ruoming},
  title     = {Conformer: Convolution-augmented transformer for speech recognition},
  booktitle = {Interspeech 2020},
  year      = {2020}
}

@article{kim2022squeeze,
  author  = {Kim, Sehoon and Kim, Dong Seuk and Vu, Thong and Kim, Jang Hoon},
  title   = {Squeezeformer: An efficient transformer for speech recognition},
  journal = {arXiv preprint arXiv:2206.00888},
  year    = {2022}
}

@inproceedings{li2023zipformer,
  author    = {Li, Z. and Wu, L. and Li, J. and Zhang, P. and Liu, X. and Kang, Y. and Li, Y. and Xie, Y. and Wang, J. and Xie, L.},
  title     = {{Zipformer}: A faster and better conformer for automatic speech recognition},
  booktitle = {International Conference on Learning Representations (ICLR)},
  year      = {2023}
}

@inproceedings{dai2019transformerxl,
  author    = {Dai, Zihang and Yang, Zhilin and Yang, Yiming and Carbonell, Jaime and Le, Quoc V and Salakhutdinov, Ruslan},
  title     = {{Transformer-XL}: Attentive language models beyond a fixed-length context},
  booktitle = {Proceedings of the 57th Annual Meeting of the Association for Computational Linguistics},
  pages     = {2978--2988},
  year      = {2019}
}

\end{document}